\documentclass{article}
\usepackage{ijcai23}

\usepackage{times}
\usepackage{soul}
\usepackage{url}
\usepackage[hidelinks]{hyperref}
\usepackage[utf8]{inputenc}
\usepackage[small]{caption}
\usepackage{graphicx}
\usepackage{amsmath}
\usepackage{amsthm}
\usepackage{booktabs}
\usepackage[switch]{lineno}
\usepackage[utf8]{inputenc}             
\usepackage{url}                        
\usepackage{booktabs}                   
\usepackage{amsfonts}                   
\usepackage{nicefrac}                   
\usepackage{microtype}                  
\usepackage{verbatim}                   
\usepackage[switch]{lineno}
\usepackage{lipsum}

\hypersetup{
   breaklinks=true
}

\usepackage{glossaries}
\glsdisablehyper
\newglossaryentry{co}{
    name=\emph{constrained optimization},
    description={},
}
\newacronym{map}{MPE}{\emph{most probable explanation}}
\newacronym{dfa}{DFSA}{Deterministic Finite State Automaton}
\newacronym{dafsa}{DAFSA}{Deterministic Acyclic Finite State Automata}
\newacronym{smp}{SMP}{\emph{structured message passing}}
\newacronym{lpi}{LPI}{\emph{Lifted probabilistic inference}}
\newacronym{wcsp}{WCSP}{Weighted Constraint Satisfaction Problem}
\newacronym{be}{BE}{Bucket Elimination}
\newacronym{mbe}{MBE}{Mini-Bucket Elimination}
\newacronym{fabe}{FABE}{``Finite state Automata Bucket Elimination''}
\newacronym{aobb}{AOBB}{\emph{branch-and-bound}}
\newacronym{aobf}{AOBF}{\emph{best-first}}
\newacronym{rbfaoo}{RBFAOO}{\emph{recursive best-first}}
\newacronym{sprbfaoo}{SPRBFAOO}{\emph{parallel recursive best-first}}
\newacronym{bdd}{BDD}{Binary Decision Diagram}
\newacronym{psdd}{PSDD}{Probabilistic Sentential Decision Diagram}
\newacronym{add}{ADD}{Algebraic Decision Diagram}
\newacronym{mdd}{MDD}{Multi-valued Decision Diagram}
\newacronym{vdd}{VDD}{Valued Decision Diagram}

\newcommand{\etal}{\emph{et al}.}

\newcommand{\wrt}{\emph{wrt}}
\newcommand{\authorref}[2]{#1~[\citeyear{#2}]}
\newcommand{\prob}[3]{\texttt{#1}{$\cdot$}\texttt{#2}{$\cdot$}\texttt{#3}}

\usepackage{pgfplots}
\pgfplotsset{
    compat=newest,
    scaled x ticks=false,
}

\usepackage{multicol}
\usepackage{xcolor,colortbl}
\usepackage{etoolbox,siunitx}
\robustify\bfseries

\usetikzlibrary{arrows,arrows.meta}
\usetikzlibrary{calc}
\usetikzlibrary{automata}
\usetikzlibrary[backgrounds]

\usepackage{mleftright,xparse}
\usepackage{mathtools}
\usepackage{mathdots}
\DeclareMathOperator*{\argmax}{arg\,max}
\DeclareMathOperator*{\argmin}{arg\,min}
\DeclareMathOperator*{\Downarrowmath}{\Downarrow}

\NewDocumentCommand\xDeclarePairedDelimiter{mmm}
 {%
  \NewDocumentCommand#1{som}{%
   \IfNoValueTF{##2}
    {\IfBooleanTF{##1}{#2##3#3}{\mleft#2##3\mright#3}}
    {\mathopen{##2#2}##3\mathclose{##2#3}}%
  }%
 }

\xDeclarePairedDelimiter{\pa}{(}{)}

\xDeclarePairedDelimiter{\interval}{[}{]}
\xDeclarePairedDelimiter{\card}{\vert}{\vert}
\xDeclarePairedDelimiter{\norm}{\lVert}{\rVert}
\xDeclarePairedDelimiter{\tuple}{\langle}{\rangle}

\newcommand{\bigO}[1]{\mathcal O\pa{#1}}
\newcommand\suchthat{\;\ifnum\currentgrouptype=16 \middle\fi|\;}

\newtheorem{definition}{Definition}
\newtheorem{remark}{Remark}

\usepackage[noend]{algpseudocode}
\usepackage[noend]{algcompatible}
\usepackage{algorithm}
\algloopdefx{RETURN}[1][]{\textbf{return} #1}

\algnewcommand{\LineComment}[1]{\State \textcolor{blue}{\{#1\}}}

\algtext*{ENDFAP}

\makeatletter
\let\OldStatex\Statex
\renewcommand{\Statex}[1][3]{%
\setlength\@tempdima{\algorithmicindent}%
\OldStatex\hskip\dimexpr#1\@tempdima\relax}
\makeatother

\title{Faster Exact \acrshort{map} and Constrained Optimization\\with Deterministic Finite State Automata}

\author{
    Filippo Bistaffa
    \affiliations
    IIIA-CSIC
    \emails
    filippo.bistaffa@iiia.csic.es
}

\begin{document}

\maketitle

\glsunset{fabe}
\begin{abstract}
We propose a concise function representation based on deterministic finite state automata for exact \acrlong{map} and \emph{constrained optimization} tasks in graphical models.
We then exploit our concise representation within \gls{be}. We denote our version of \gls{be} as \acrshort{fabe}.
\gls{fabe} significantly improves the performance of \gls{be} in terms of runtime and memory requirements by minimizing redundancy.
Results on \acrlong{map} and \emph{weighted constraint satisfaction} benchmarks show that \acrshort{fabe} often outperforms the state of the art, leading to significant runtime improvements (up to 5 orders of magnitude in our tests). 
\end{abstract}

\glsreset{be}
\section{Introduction}

Graphical models are a widely used theoretical framework that provides the basis for many reasoning tasks both with probabilistic and deterministic information~\cite{DBLP:series/synthesis/2013Dechter}.
These models employ graphs to concisely represent the structure of the problem and the relations among variables.

One of the most important algorithms for exact reasoning on graphical models is \gls{be} proposed by \authorref{Dechter}{DBLP:conf/uai/Dechter96,DBLP:series/synthesis/2013Dechter},
a general approach based on the concept of \emph{variable elimination} that accommodates many inference and optimization tasks, including \gls{map} and \emph{constrained optimization}.
Solving these tasks is fundamental in real-world applications domains such as genetic linkage analysis \cite{fishelson2002exact}, protein side-chain interaction \cite{yanover2008minimizing}, and Earth observation satellite management \cite{bensana1999earth}. 
\gls{be} is also a fundamental component---the approximate version of \gls{be} \cite{DBLP:conf/ijcai/Dechter97} is used to compute the initial heuristic that guides the search---of all the AND/OR search algorithms by \authorref{Marinescu \etal}{DBLP:conf/aaai/MarinescuD07,DBLP:journals/ai/MarinescuD09,DBLP:conf/uai/Kishomoto014,DBLP:conf/nips/KishimotoMB15} that represent the state of the art for exact \gls{map} inference.
On the other hand, \gls{be} is characterized by memory requirements that grow \emph{exponentially} with respect to the \emph{induced width} of the primal graph associated to the graphical model \cite{DBLP:series/synthesis/2013Dechter}, severely hindering its applicability to large exact reasoning tasks.
As a consequence, several works have tried to mitigate this drawback \cite{DBLP:conf/ijcai/Dechter97,DBLP:journals/tcyb/BistaffaBF17,bistaffa2018cop}, but none of these approaches really managed to overcome such a limitation.
One of the major reasons for such memory requirements is the fact that the functions employed during \gls{be}'s execution are usually represented as \emph{tables}, whose size is the product of the domains of the variables in the scope, regardless of the actual values of such functions.
This representation can lead to storing many repeated values in the same table, causing a potential waste of memory and other computational resources. 

Against this background, in this paper we propose a function representation specifically devised for exact \gls{map} inference and constrained optimization that, instead of the traditional mapping \emph{variable assignment} $\to$ \emph{value}, adopts a different approach that maps each value $v$ to the minimal \emph{finite state automaton} \cite{DBLP:books/aw/HopcroftU79} representing all the variable assignments that are associated to $v$.

Compact function representations have already been investigated by different works \cite{srinivasan1990algorithms,bahar1997algebric,chavira2007compiling,10.5555/3023638.3023664}, which have analyzed the potential benefits of the use of different types of \emph{decision diagrams} from a theoretical point of view.
Nonetheless, as noted by \authorref{Chavira and Darwiche}{chavira2007compiling} the effectiveness of these approaches has been limited in practice, because the overhead they incur may very well outweigh any gains.
Indeed, no approach based on decision diagrams has ever been shown to outperform the current state of the art, i.e., the AND/OR search algorithms by \authorref{Marinescu \etal}{DBLP:conf/uai/Kishomoto014,DBLP:conf/nips/KishimotoMB15}.

To address the above-mentioned limitations, in this paper we propose the novel use of automata theory techniques to efficiently represent and manipulate functions in a compact way within our approach called \acrshort{fabe}.
By representing each value only once, and by exploiting the well-known capabilities of automata of compactly representing sets of strings (with a reduction that can be \emph{up to exponential} with respect to a full table), we significantly improve the performance of \gls{be} in terms of runtime and memory requirements.
In more detail, this paper advances the state of the art in the following ways:
\begin{itemize}
\item We propose a function representation for exact \gls{map} inference and constrained optimization based on finite state automata, which we exploit within \acrshort{fabe}.
\item Results on standard benchmark datasets show that \acrshort{fabe} often outperforms the current state of the art, with improvements of up to 5 orders of magnitude in our tests.
\item Results also show that \acrshort{fabe} outperforms the \gls{smp} approach by \authorref{Gogate and Domingos}{10.5555/3023638.3023664}, i.e., the most recent approach in the literature based on decision diagrams.
\item Our work paves the way for an improved version of \gls{be} as a key component of AND/OR search algorithms, in which the computation of the initial heuristic can represent a bottleneck \cite{DBLP:conf/nips/KishimotoMB15}.
\end{itemize}

This paper is structured as follows.
Section~\ref{sec:back} provides the background on graphical models and deterministic finite state automata.
Section~\ref{sec:relwork} positions our approach \wrt{} existing literature.
Section~\ref{sec:dafsa} presents our function representation and how we exploit it within \gls{fabe}.
Section~\ref{sec:exp} presents our experimental evaluation on standard benchmark datasets, in which we compare \gls{fabe} against the state of the art.
Section~\ref{sec:concl} concludes the paper and outlines future research directions.

\section{Background}
\label{sec:back}

\subsection{Graphical Models}

\emph{Graphical models} (e.g., Bayesian Networks \cite{DBLP:books/daglib/0066829}, Markov Random Fields \cite{lauritzen1996graphical}, and Cost Networks \cite{DBLP:series/synthesis/2013Dechter}) capture the factorization structure of a distribution over a set of $n$ variables.
Formally, a graphical model is a tuple $\mathcal M = \tuple{\mathbf X, \mathbf D, \mathbf F}$, where $\mathbf X = \{ X_i : i \in V\}$ is a set of variables indexed by set $V$ and $\mathbf D = \{ D_i : i \in V\}$ is the set of their finite domains of values.
$\mathbf F = \{ \psi_\alpha : \alpha \in F\}$ is a set of discrete local functions defined on subsets of variables, where $F \subseteq 2^V$ is a set of variable subsets.
We use $\alpha \subseteq V$ and $\mathbf X_\alpha \subseteq \mathbf X$ to indicate the \emph{scope} of function $\psi_\alpha$, i.e., $\mathbf X_\alpha = var(\psi_\alpha) = \{X_i : i \in \alpha\}$.
The function scopes yield a \emph{primal graph} $G$ whose vertices are the variables and whose edges connect any two variables that appear in the scope of the same function.
An important inference task that appears in many real-world applications is \gls{map}. \gls{map} finds a complete assignment to the variables that has the highest probability (i.e., a mode of the joint probability), namely: $\mathbf x^*=\argmax_{\mathbf x}\prod_{\alpha\in F}\psi_\alpha(\mathbf X_\alpha)$.
The task is NP-hard \cite{DBLP:books/daglib/0066829}.
Another important task over deterministic graphical models (e.g., Cost Networks) is the optimization task of finding an assignment or a configuration to all the variables that minimizes the sum of the local functions, namely: $\mathbf x^*=\argmin_{\mathbf x}\sum_{\alpha\in F}\psi_\alpha(\mathbf X_\alpha)$.
This is the task that has to be solved in \glspl{wcsp}.
The task is NP-hard to solve \cite{DBLP:series/synthesis/2013Dechter}.

To solve the above-mentioned tasks we consider the \gls{be} algorithm as discussed by \authorref{Dechter}{DBLP:series/synthesis/2013Dechter} (Algorithm \ref{alg:be}).
\gls{be} is a general algorithm that can accommodate several exact inference and optimization tasks over graphical models.
In this paper we focus on the version that can optimally solve the above-mentioned \gls{map} and optimization tasks.
\gls{be} operates on the basis of a \emph{variable ordering} $d$, which is used to partition the set of functions into sets called \emph{buckets}, each associated with one variable of the graphical model.
Each function is placed in the bucket associated with the last bucket that is associated with a variable in its scope.
Then, buckets are processed from last to first by means of two fundamental operations, i.e., \emph{combination} ($\otimes\in\{\prod,\sum\}$) and \emph{projection} ($\Downarrowmath\in\{\max,\min\}$).
All the functions in $bucket_p$, i.e., the current bucket, are combined with the $\otimes$ operation, and the result is the input of a $\Downarrowmath$ operation.
Such an operation removes $X_p$ from the scope, producing a new function $h_p$ that does not involve $X_p$, which is then placed in the last bucket that is associated with a variable appearing in the scope of the new function.
To solve the \gls{map} (resp. optimization) task, $\otimes=\prod$ (resp. $\sum$) and $\Downarrowmath=\max$ (resp. $\min$) operators are used.

\begin{algorithm}[t]\caption{\Acrlong{be} \protect\cite{DBLP:series/synthesis/2013Dechter}}\label{alg:be}
\begin{algorithmic}[1]
\REQUIRE A graphical model $\mathcal M = \tuple{\mathbf X, \mathbf D, \mathbf F}$, an ordering $d$.
\ENSURE A max probability (resp. min cost) assignment.
\STATE Partition functions into buckets according to $d$.
\STATE Define $\psi_i$ as the $\otimes$ of $bucket_i$ associated with $X_i$.
\FOR{$p\gets n$ down to 1}
\FOR{$\psi_p$ and messages $h_1,h_2,\ldots,h_j$ in $bucket_p$}
\STATE $h_p\gets\Downarrowmath_{X_p}(\psi_p\otimes\bigotimes_{i=1}^j h_i)$.
\STATE Place $h_p$ into the largest index variable in its scope.
\ENDFOR
\ENDFOR
\STATE Assign maximizing (resp. minimizing) values in ordering $d$, consulting functions in each bucket.
\RETURN Optimal solution value and assignment.
\end{algorithmic}
\end{algorithm}

The computational complexity of the \gls{be} algorithm is directly determined by the ordering $d$.
Formally, \gls{be}'s time and space complexity are $\bigO{r\cdot k^{w^*(d)+1}}$ and $\bigO{n\cdot k^{w^*(d)}}$ respectively, where $k$ bounds the domain size, and $w^*(d)$ is the induced width of its primal graph along $d$ \cite{DBLP:series/synthesis/2013Dechter}.
Hence, it is of utmost importance to adopt a variable ordering $d$ that minimizes the induced width $w^*(d)$.
The task of computing such ordering is NP-complete, and, for this reason, a greedy procedure is usually adopted to compute a variable ordering of acceptable quality.
Several $metric(\cdot)$ functions have been proposed in the literature.
One of the most widely used is \textsc{min-fill}, such that $metric(X_i)$ is the number of fill edges for $X_i$'s parents in the primal graph \cite{DBLP:series/synthesis/2013Dechter}.

\subsection{Deterministic Finite State Automata}

Let $\Sigma$ denote a finite alphabet of characters and $\Sigma^*$ denote the set of all strings over $\Sigma$.
The size $\card{\Sigma}$ of $\Sigma$ is the number of characters in $\Sigma$.
A language over $\Sigma$ is any subset of $\Sigma^*$.
A \gls{dfa} \cite{DBLP:books/aw/HopcroftU79} $\delta$ is specified by a tuple $\tuple{Q,\Sigma,t,s,F}$, where $Q$ is a finite set of states, $\Sigma$ is an input alphabet, $t : Q \times\Sigma\to 2^Q$ is a transition function, $s\in Q$ is the start state and $F \subseteq Q$ is a set of final states. 
A string $x$ over $\Sigma$ is accepted (or recognized) by $\delta$ if there is a labeled path from $s$ to a final state in $F$ such that this path spells out the string $x$.
Thus, the language $L_\delta$ of a \gls{dfa} $\delta$ is the set of all strings that are spelled out by paths from $s$ to a final state in $F$.
It is well known that a general \gls{dfa} can accept an \emph{infinite} language (i.e., a infinite set of strings) \cite{DBLP:books/aw/HopcroftU79}.
In this paper we focus on \gls{dafsa}, i.e., \gls{dfa} whose corresponding graph is a directed acyclic graph.
In contrast with general \gls{dfa}, \gls{dafsa} only accept \emph{finite} languages \cite{daciuk2002comparison}.

\section{Related Work}\label{sec:relwork}

In recent years, a strand of literature has investigated the use of different algorithms on AND/OR search spaces, progressively showing the effectiveness of these approaches for exact \gls{map} inference and constrained optimization.
Specifically, \gls{aobb} \cite{DBLP:journals/ai/MarinescuD09}, \gls{aobf} \cite{DBLP:conf/aaai/MarinescuD07}, \gls{rbfaoo} \cite{DBLP:conf/uai/Kishomoto014} and \gls{sprbfaoo} \cite{DBLP:conf/nips/KishimotoMB15} algorithms have been proposed.
To this day, \gls{rbfaoo} \cite{DBLP:conf/uai/Kishomoto014} and \gls{sprbfaoo} \cite{DBLP:conf/nips/KishimotoMB15} still represent the state of the art for exact \gls{map} inference, since, to the best of our knowledge, no algorithms have been shown to outperform them in terms of runtime.
For this reason, we compare our approach against these two algorithms in Section \ref{sec:exp}.
All these approaches use the standard tabular representation to store functions in memory.
In the context of constrained optimization, the only approach that tries to reduce the size of tables in memory is the one by \authorref{Bistaffa \etal}{DBLP:journals/tcyb/BistaffaBF17}, which avoids representing \emph{unfeasible} assignments during the solution of \glspl{wcsp} with \gls{be}.

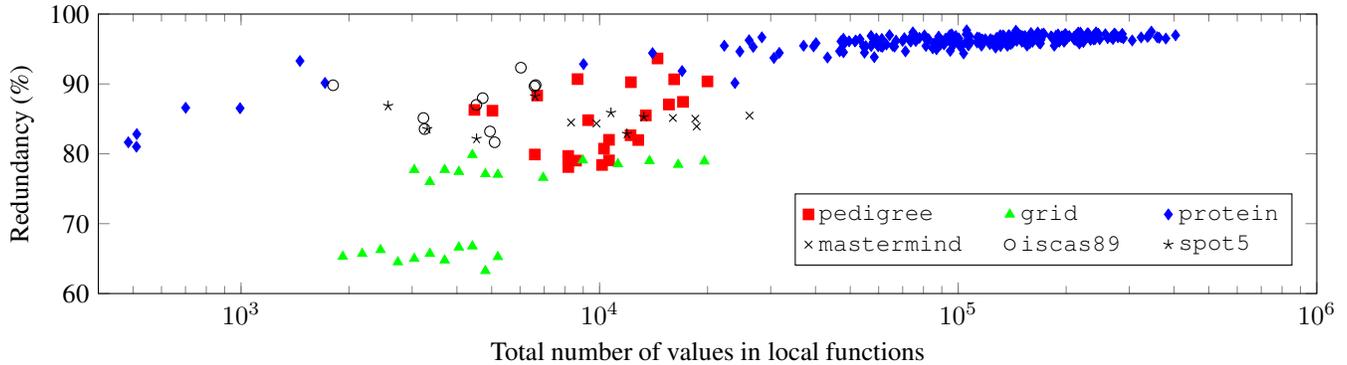
\begin{figure*}[t]
\centering
\begin{tikzpicture}
\begin{axis}[
    width=\textwidth,
    height=5.3cm,
    ymin=0.6,
    ymax=1,
    yticklabels={50,60,70,80,90,100},
    xmin=400,
    xmax=1000000,
    xmode=log,
    xlabel={Total number of values in local functions},
    ylabel={Redundancy (\%)},
    y label style={at={(axis description cs:-0.08,0.5)},anchor=north},
    legend entries={
        \texttt{pedigree},
        \texttt{grid},
        \texttt{protein},
        \texttt{mastermind},
        \texttt{iscas89},
        \texttt{spot5},
    },
    legend cell align={left},
    legend style={
        font=\small,
        at={(0.98,0.1)},
        anchor=south east,
        /tikz/every even column/.append style={column sep=5mm}
    },
    legend columns=3,
]
    \addplot[
        scatter/classes={
            pedigree={mark=square*,red},
            grid={mark=triangle*,green},
            protein={mark=diamond*,blue},
            mastermind={mark=x},
            iscas89={mark=o},
            spot5={mark=star}
        },
        scatter, mark=*, only marks, 
        scatter src=explicit symbolic,
    ] table [meta=class] {
        x y class
        13456 0.854935 pedigree
        12819 0.819721 pedigree
        6597 0.799 pedigree
        8680 0.906912 pedigree
        4476 0.862824 pedigree
        8174 0.796917 pedigree
        15613 0.870557 pedigree
        17077 0.874334 pedigree
        10285 0.807389 pedigree
        16143 0.906647 pedigree
        12198 0.826529 pedigree
        5025 0.861692 pedigree
        10622 0.790623 pedigree
        10622 0.819996 pedigree
        8179 0.781025 pedigree
        6698 0.883398 pedigree
        8596 0.790251 pedigree
        10158 0.784013 pedigree
        12225 0.902495 pedigree
        19986 0.903633 pedigree
        9290 0.848116 pedigree
        14505 0.936436 pedigree
        19602 0.789307 grid
        4418 0.798551 grid
        1682 0.433413 grid
        3698 0.647647 grid
        2738 0.417823 grid
        3362 0.657049 grid
        3362 0.759964 grid
        13778 0.789882 grid
        8978 0.790822 grid
        6962 0.765872 grid
        2738 0.644996 grid
        5202 0.652441 grid
        4802 0.771137 grid
        3042 0.412229 grid
        3698 0.777177 grid
        2450 0.41102 grid
        1922 0.652966 grid
        4418 0.667723 grid
        2178 0.414141 grid
        2178 0.657025 grid
        3042 0.649901 grid
        4050 0.666173 grid
        1922 0.424037 grid
        1058 0.372401 grid
        16562 0.784446 grid
        4050 0.774321 grid
        5202 0.770281 grid
        3042 0.77712 grid
        1458 0.40535 grid
        4802 0.632445 grid
        11250 0.785422 grid
        2450 0.662449 grid
        76171 0.961534 protein
        94759 0.969839 protein
        47843 0.964593 protein
        149644 0.96914 protein
        269468 0.968286 protein
        166832 0.971067 protein
        230619 0.972184 protein
        24611 0.946406 protein
        242328 0.973812 protein
        379428 0.964974 protein
        148504 0.966412 protein
        89242 0.95631 protein
        63101 0.956419 protein
        238792 0.963592 protein
        339339 0.968677 protein
        102293 0.954308 protein
        122516 0.957042 protein
        82638 0.96423 protein
        199310 0.971316 protein
        173260 0.971967 protein
        116761 0.96349 protein
        47881 0.945385 protein
        100432 0.965051 protein
        98298 0.962766 protein
        61244 0.964454 protein
        133325 0.969413 protein
        282985 0.967935 protein
        229025 0.967523 protein
        286768 0.971601 protein
        73714 0.960604 protein
        235420 0.965555 protein
        127957 0.96591 protein
        240845 0.964388 protein
        53236 0.964028 protein
        161195 0.960315 protein
        74745 0.973697 protein
        129754 0.962899 protein
        136371 0.95744 protein
        131009 0.969697 protein
        403850 0.969707 protein
        98478 0.968541 protein
        95927 0.970905 protein
        59127 0.965024 protein
        108100 0.960527 protein
        69988 0.961608 protein
        143410 0.97095 protein
        228623 0.967899 protein
        345847 0.974789 protein
        26817 0.953015 protein
        139517 0.960485 protein
        47903 0.956433 protein
        511 0.810176 protein
        90441 0.961168 protein
        270111 0.964315 protein
        233931 0.963066 protein
        169541 0.957863 protein
        101145 0.954125 protein
        105604 0.976734 protein
        139087 0.968861 protein
        994 0.865191 protein
        30631 0.93722 protein
        154146 0.964683 protein
        229218 0.972349 protein
        66457 0.953459 protein
        124618 0.962501 protein
        61452 0.953199 protein
        162699 0.966238 protein
        106816 0.955166 protein
        1715 0.901458 protein
        113495 0.961161 protein
        89556 0.961923 protein
        80526 0.969476 protein
        168254 0.958598 protein
        47105 0.96096 protein
        26115 0.96255 protein
        140615 0.964101 protein
        181515 0.967937 protein
        159611 0.972884 protein
        138789 0.964255 protein
        187368 0.968981 protein
        212557 0.968822 protein
        14050 0.944128 protein
        212891 0.969369 protein
        323754 0.966626 protein
        40131 0.958312 protein
        43203 0.937782 protein
        115359 0.959942 protein
        54795 0.944758 protein
        261115 0.965575 protein
        138663 0.966271 protein
        337262 0.967367 protein
        57202 0.954215 protein
        49629 0.961091 protein
        96226 0.957288 protein
        200807 0.965071 protein
        701 0.865906 protein
        363410 0.966104 protein
        105626 0.959678 protein
        170230 0.963414 protein
        28349 0.966701 protein
        138335 0.969783 protein
        46713 0.946803 protein
        153648 0.968493 protein
        118991 0.963938 protein
        105278 0.971001 protein
        9020 0.928381 protein
        210673 0.96364 protein
        87906 0.962426 protein
        212362 0.964104 protein
        143502 0.968356 protein
        57879 0.959536 protein
        184714 0.964009 protein
        37056 0.954609 protein
        70779 0.96547 protein
        92237 0.964797 protein
        267142 0.967033 protein
        221632 0.965722 protein
        194766 0.968336 protein
        152973 0.961542 protein
        88209 0.96769 protein
        248107 0.96412 protein
        114631 0.965725 protein
        148935 0.966247 protein
        134880 0.963983 protein
        87014 0.949721 protein
        278865 0.966887 protein
        191391 0.960458 protein
        198633 0.962599 protein
        158448 0.958485 protein
        81599 0.963149 protein
        219192 0.972075 protein
        173155 0.97046 protein
        189053 0.966084 protein
        171329 0.9714 protein
        197359 0.960468 protein
        103531 0.943708 protein
        169510 0.968332 protein
        305663 0.962243 protein
        139695 0.963764 protein
        51538 0.960844 protein
        219550 0.966759 protein
        99693 0.961733 protein
        142087 0.964191 protein
        246097 0.971743 protein
        39512 0.953786 protein
        189888 0.970778 protein
        95026 0.94732 protein
        113011 0.96641 protein
        150665 0.960422 protein
        167850 0.965999 protein
        145689 0.961823 protein
        272557 0.972006 protein
        54435 0.95378 protein
        112881 0.959347 protein
        75500 0.965748 protein
        185348 0.954734 protein
        512 0.828125 protein
        132719 0.967684 protein
        1460 0.932877 protein
        16984 0.918453 protein
        274293 0.971078 protein
        84927 0.951005 protein
        97134 0.9564 protein
        75985 0.962664 protein
        219195 0.973403 protein
        60887 0.954046 protein
        81273 0.954376 protein
        65681 0.954584 protein
        85519 0.957659 protein
        288866 0.965472 protein
        137199 0.954584 protein
        285886 0.967102 protein
        71281 0.946648 protein
        146600 0.972906 protein
        217750 0.957814 protein
        213650 0.96227 protein
        107160 0.969149 protein
        86045 0.956232 protein
        75924 0.969246 protein
        187221 0.95876 protein
        114528 0.965161 protein
        51813 0.962037 protein
        223436 0.960015 protein
        214000 0.968565 protein
        163123 0.965278 protein
        358895 0.966898 protein
        22270 0.954648 protein
        73107 0.963492 protein
        159248 0.968364 protein
        93707 0.963311 protein
        134207 0.968251 protein
        340974 0.970523 protein
        23844 0.901359 protein
        31706 0.944427 protein
        189589 0.961485 protein
        105828 0.961957 protein
        182348 0.967398 protein
        123036 0.964059 protein
        240842 0.967331 protein
        59124 0.958579 protein
        151855 0.962919 protein
        77974 0.965989 protein
        210429 0.967533 protein
        231592 0.969192 protein
        238148 0.9671 protein
        216028 0.973429 protein
        96218 0.960714 protein
        55232 0.954121 protein
        59824 0.953982 protein
        144844 0.976098 protein
        127121 0.952069 protein
        155603 0.95493 protein
        58347 0.93842 protein
        190048 0.962867 protein
        131229 0.954256 protein
        64343 0.969787 protein
        89755 0.961072 protein
        198258 0.970589 protein
        238398 0.969501 protein
        107461 0.968286 protein
        485 0.816495 protein
        91306 0.956596 protein
        129458 0.960157 protein
        107864 0.959217 protein
        189673 0.962335 protein
        57147 0.968677 protein
        211204 0.964082 protein
        153386 0.963745 protein
        257245 0.971156 protein
        116136 0.954812 protein
        67093 0.961859 protein
        151699 0.967996 protein
        184727 0.964266 protein
        159537 0.961777 protein
        204037 0.968555 protein
        285226 0.967229 protein
        141746 0.959032 protein
        165968 0.963523 protein
        140040 0.957105 protein
        166660 0.973629 protein
        26186 0.854655 mastermind
        8326 0.844823 mastermind
        18488 0.850173 mastermind
        18658 0.839104 mastermind
        9802 0.843399 mastermind
        16008 0.851262 mastermind
        6632 0.898372 iscas89
        6032 0.923243 iscas89
        5096 0.816523 iscas89
        3226 0.851209 iscas89
        1806 0.898117 iscas89
        4532 0.870035 iscas89
        4946 0.831985 iscas89
        4720 0.879661 iscas89
        6584 0.896719 iscas89
        3250 0.835385 iscas89
        3306 0.835451 spot5
        6606 0.882228 spot5
        4538 0.821507 spot5
        13288 0.8528 spot5
        2572 0.868585 spot5
        10758 0.85871 spot5
        11920 0.828607 spot5
	};
\end{axis}
\end{tikzpicture}
\caption{\label{fig:red}Redundancy in \gls{map} and \glspl{wcsp} instances. Best viewed in colors.}
\end{figure*}

The task of concisely representing functions for inference has been studied by different works over the years.
In the context of this work, the most closely related strand of literature is concerned with the use of \emph{decision diagrams} to compactly represent knowledge.
A decision diagram is a directed acyclic graph that maps several decision nodes to two or more terminal nodes.
Different types of decision diagrams exist, depending on the domains of the variables and the types of employed operations.
Along these lines, several works have proposed the use of \glspl{bdd} \cite{10.5555/3023638.3023664}, \glspl{add} \cite{bahar1997algebric}, affine \glspl{add} \cite{10.5555/1642293.1642513}, \glspl{mdd} \cite{srinivasan1990algorithms,chavira2007compiling} within variable elimination and/or message passing algorithms for inference.

While it is true that \gls{dafsa} and decision diagrams are very closely related, to the best of our knowledge the use of automata theory has never been investigated in the context of concise function representation and inference.
In this sense, the use of \gls{dafsa} cannot be considered entirely novel (indeed, a \gls{dafsa} can be considered equivalent to an \gls{mdd}).
On the other hand, our proposed use of automata theory techniques for compilation \cite{daciuk2002comparison}, minimization \cite{DBLP:conf/stringology/Bubenzer11}, union \cite{han2008state}, and determinization \cite{DBLP:books/aw/HopcroftU79} is, as far as we know, entirely novel in the context of inference on graphical models.

Moreover, as noted by \authorref{Chavira and Darwiche}{chavira2007compiling} the effectiveness of approaches based on decision diagrams has been limited in practice, because the overhead they incur may very well outweigh any gains.
Indeed, no approach based on any compact representations has ever been shown to outperform the current state of the art, i.e., the AND/OR search algorithms by \authorref{Marinescu \etal}{DBLP:conf/uai/Kishomoto014,DBLP:conf/nips/KishimotoMB15}.
Against this background, we believe that our contribution, i.e., to show that a new approach based on well-known theoretical concepts can still be competitive with the current state of the art, can be very useful for the scientific community.

Given the limited amount of available space, in this paper we decided to compare \gls{fabe} only with the \gls{smp} approach by \authorref{Gogate and Domingos}{10.5555/3023638.3023664}, since it is, to the best of our knowledge, the most recent and relevant one among the above-mentioned works focusing on decision diagrams.\footnote{Our decision is also motivated by the fact that, to the best of our knowledge, none of these works has an available implementation.}
We also remark that, despite the importance of considering this strand of literature, the current state of the art is represented by more recent approaches based on completely different techniques \cite{DBLP:conf/uai/Kishomoto014,DBLP:conf/nips/KishimotoMB15}.
Hence, our evaluation in Section \ref{sec:exp} is mainly devoted to the comparison against these approaches.


The use of \glspl{mdd} has also been investigated by \authorref{Mateescu \etal}{mateescu2008and} within the above-mentioned AND/OR search algorithms, but this approach has been subsumed and outperformed by more recent and advanced approaches based on AND/OR search trees, such as (SP)\gls{rbfaoo}.
For this reason, in Section \ref{sec:exp} we only compare with the most recent ones in such a strand of literature.\footnote{We cannot directly compare with the approach by \authorref{Mateescu \etal}{mateescu2008and} also because its implementation is not publicly available.}

\gls{lpi} \cite{kersting2012lifted} is also concerned with reducing redundancy within probabilistic inference.
Specifically, \gls{lpi} tackles redundancy \emph{between different factors}, whereas we tackle redundancy \emph{inside the same factor}.
Assessing the effectiveness of the combined approach \wrt{} to the separate ones is a non-trivial research question, which will be considered in future work.
Finally, the approach by \authorref{Demeulenaere \etal}{demeulenaere2016compact} aims at representing \emph{compact tables} by maintaining \emph{generalized arc consistency} (i.e., by removing from domains all values that have no support on a constraint) and not by minimizing redundancy due to repeated values, as we explain in the next section.

\def\dafsa{D}

\section{A Novel \gls{dafsa}-based Version of \gls{be}}
\label{sec:dafsa}

All the datasets commonly used as benchmarks for \gls{map} \cite{DBLP:conf/nips/KishimotoMB15} 
and constrained optimization 
\cite{DBLP:journals/ai/MarinescuD09} are characterized by a very high \emph{redundancy}, i.e., many different variable assignments are associated to the same value in the local functions.
Figure~\ref{fig:red} shows that the value of redundancy for local functions (defined as $1-\frac{\text{number of unique values}}{\text{total number of values}}$) for all \gls{map} and \gls{wcsp} instances is always above 80\% (except for smaller \texttt{grid} instances).

Furthermore, in probabilistic graphical models, local functions represent probabilities with values in the interval $[0,1]$, which, in theory, contains \emph{infinite} real values.
In practice, such values are represented by \emph{floating point numbers} that can only represent a \emph{finite} amount of values.
Thus, while a table $\psi$ has an arbitrarily large size that is the product of the domains of the variables in its scope, in practice the maximum number of unique values in $\psi$ is bounded by a parameter that depends on the numerical representation.
These remarks motivate the study of a novel concise representation that exploits such a redundancy to reduce the amount of computation.
Notice that state of the art approaches for exact inference~\cite{DBLP:conf/nips/KishimotoMB15} represent functions as \emph{full tables}, whose size is the product of the domains of the variables in the scope. 

\begin{figure}[b]
\centering
\begin{tikzpicture}[yscale=0.71,xscale=0.56]

\tikzstyle{dafsastate}=[state,inner sep=0,minimum width=3mm,minimum height=3mm]
\def\v{6mm}

\node[fill=blue!20!white,draw,minimum width=\v,minimum height=\v] at (0,0) (b) {$v_2$};
\node[draw,minimum width=\v,minimum height=\v] at ($ (b) + (0,1.5) $) (a) {$v_1$};
\node[fill=green!20!white,draw,minimum width=\v,minimum height=\v] at ($ (b) - (0,1.5) $) (c) {$v_3$};
\node[minimum width=\v] at ($ (a) + (0,1.25) $) (psi) {$\psi$};


\node[dafsastate] at ($(a) + (2,0)$) (a1) {};
\node[dafsastate] at ($(a1) + (2,0)$) (a2) {};
\node[dafsastate] at ($(a2) + (2,0.25)$) (a3a) {};
\node[dafsastate] at ($(a2) + (2,-0.25)$) (a3b) {};
\node[dafsastate,accepting] at ($(a2) + (4,0)$) (a4) {};

\path[-{Stealth}]
(a) edge [above] node [align=center] {} (a1)
(a1) edge [above] node [align=center] {0} (a2)
(a2) edge [above,sloped] node [align=center] {0} (a3a)
(a3a) edge [above,sloped] node [align=center] {$\ast$} (a4);

\path[-{Stealth}]
(a2) edge [below,sloped] node [align=center] {1} (a3b)
(a3b) edge [below,sloped] node [align=center] {1} (a4);


\node[dafsastate] at ($(b) + (2,0)$) (b1) {};
\node[dafsastate] at ($(b1) + (2,0)$) (b2) {};
\node[dafsastate] at ($(b2) + (2,0)$) (b3) {};
\node[dafsastate,accepting] at ($(b3) + (2,0)$) (b4) {};

\path[-{Stealth}]
(b) edge [above] node [align=center] {} (b1)
(b1) edge [above] node [align=center] {0} (b2)
(b2) edge [above] node [align=center] {1} (b3)
(b3) edge [above] node [align=center] {0} (b4);


\node[dafsastate] at ($(c) + (2,0)$) (c1) {};
\node[dafsastate] at ($(c1) + (2,0)$) (c2) {};
\node[dafsastate] at ($(c2) + (2,0)$) (c3) {};
\node[dafsastate,accepting] at ($(c3) + (2,0)$) (c4) {};

\path[-{Stealth}]
(c) edge [above] node [align=center] {} (c1)
(c1) edge [above] node [align=center] {1} (c2)
(c2) edge [above] node [align=center] {$\ast$} (c3)
(c3) edge [above] node [align=center] {0} (c4);


\node at ($(b1)!0.5!(b2)$) (x1) {};
\node at (x1 |- psi) (X1) {$X_1$};

\node at ($(b2)!0.5!(b3)$) (x2) {};
\node at (x2 |- psi) (X2) {$X_2$};

\node at ($(b3)!0.5!(b4)$) (x4) {};
\node at (x4 |- psi) (X4) {$X_4$};



\node[inner sep=0,outer sep=0] at ($(psi.north)!0.5!(c.south) - (5,0)$) (table) {%
\begin{tabular}{ccc|c}
$X_1$ & $X_2$ & $X_4$ & $\psi$ \\ \hline
0 & 0 & 0 & $v_1$ \\
0 & 0 & 1 & $v_1$ \\
\rowcolor{blue!20!white}
0 & 1 & 0 & $v_2$ \\
0 & 1 & 1 & $v_1$ \\
\rowcolor{green!20!white}
1 & 0 & 0 & $v_3$ \\
\rowcolor{red!20!white}
1 & 0 & 1 & $\infty$ \\
\rowcolor{green!20!white}
1 & 1 & 0 & $v_3$ \\
\rowcolor{red!20!white}
1 & 1 & 1 & $\infty$
\end{tabular}};

\end{tikzpicture}
\caption{\label{fig:table_dafsa}Standard table (left) and corresponding \gls{dafsa}-based representation (right). All variables are binary. Best viewed in colors.}
\end{figure}
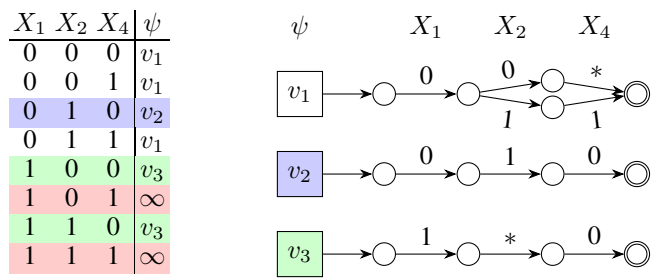

In this paper we propose a way to represent functions by means of \gls{dafsa}, as shown in the example in Figure~\ref{fig:table_dafsa}.
In the traditional way of representing functions as tables, rows are indexed using variable assignments as \emph{keys} (Figure~\ref{fig:table_dafsa}, left).

\def\reducecol{4.5pt}
\addtolength{\tabcolsep}{-\reducecol}

In contrast, here we propose a novel approach that uses \emph{values} as keys (Figure~\ref{fig:table_dafsa}, right).
Formally,
\begin{definition}
Given a function $\psi$ that maps each possible assignment of the variables in its scope to a value $v\in\mathbb{R}\cup\{\infty\}$,\footnote{We allow $\infty$ as a possible value, since it can used to represent variable assignments that violate hard constraints in \glspl{wcsp}.} we denote as $\dafsa(\psi)$ its corresponding representation in terms of \gls{dafsa}. Formally, $\dafsa(\psi)=\{(v,\delta)\}$, where $v$ is a value in $\psi$ and $\delta$ is the \emph{minimal} \gls{dafsa} that accepts all the strings corresponding to the variable assignments that were mapped to $v$ in $\psi$.
For the sake of simplicity, we do not represent the scope of the function in $\dafsa(\psi)$, as we assume it is equal to $var(\psi)$.
We label a transition that accepts all the values of a variable's domain as $\ast$.
Notice that each $\delta$ is acyclic because it accepts a finite language \cite{daciuk2002comparison}.
\end{definition}
\begin{remark}\label{rem:float}
Given that values are employed as keys in our function representation, it is crucial to ensure the absence of duplicates in such a set of keys, i.e., we must be able to correctly determine whether two values $v_1$ and $v_2$ are equal.
While this is a trivial task in theory, in practice it can be very tricky when $v_1$ and $v_2$ are floating point numbers representing real values.
Indeed, even if $v_1$ and $v_2$ are theoretically equal, their floating point representations can differ due to numerical errors implicit in floating point arithmetic, especially if $v_1$ and $v_2$ are the result of a series of operations whose numerical errors have accumulated. 
To mitigate this aspect, we use a well-known technique for comparing floating point numbers known as \emph{$\epsilon$-comparison}, i.e., $v_1$ and $v_2$ are considered equal if they differ by a quantity smaller than a small $\epsilon$.
While there exist more advanced techniques of tackling numerical issues related to floating point arithmetic~\cite{higham2002accuracy}, they are well beyond the scope of this paper.
This should \emph{not} be considered as an approximation, rather as a standard method to avoid the propagation of numerical errors.
\end{remark}

A crucial property of \gls{dafsa} is that one path can accept multiple strings, or, in our case, represent multiple variable assignments.
In the example in Figure~\ref{fig:table_dafsa}, the \gls{dafsa} corresponding to $v_3$ contains only one path, but it represents both $\tuple{1,0,0}$ and $\tuple{1,1,0}$.
By exploiting this property, our representation can reach a reduction in terms of memory that is, in the best case, \emph{up to} exponential \wrt{} the traditional table representation.
We remark that memory is the main bottleneck that limits the scalability of \gls{be}, hence reducing its memory requirements is crucial, leading to significant improvements as shown by our results in the experimental section.
Finally, our representation allows one to trivially avoid representing unfeasible assignments, similarly to \cite{DBLP:journals/tcyb/BistaffaBF17}.
Predicting the size (e.g., the number of states) of a minimal \gls{dafsa} accepting a given set of strings remains, to the best of our knowledge, an open problem, since it depends on the common prefixes/suffixes of the input set.

A minimal \gls{dafsa} can be efficiently constructed from a set of assignments by using the algorithm described by \authorref{Daciuk}{daciuk2002comparison}.
Since all the strings accepted by each \gls{dafsa} are of the same length (equal to the cardinality of the scope of the function),
so are all the paths in the \gls{dafsa}.
Thus, there is a mapping between each edge at depth $i$ in each path and the $i$\textsuperscript{th} variable in the scope (see Figure~\ref{fig:table_dafsa}).
Without loss of generality, our representation always maintains the variables in the scope ordered \wrt{} their natural ordering.
We now discuss our \gls{dafsa}-based version of \gls{be}, i.e., its $\otimes$ and $\Downarrowmath$ operations.

\subsection{A \gls{dafsa}-Based Version of $\otimes$}

In order to better discuss our \gls{dafsa}-based version the $\otimes$ operation, let us first recall how this operation works for traditional tabular functions with an example (Figure~\ref{fig:sum}).

\begin{figure}[H]
\centering
\begin{tikzpicture}[
    every node/.append style={inner sep=0,outer sep=0},
    xscale=0.62,
]

\node at (0,0) (t1) {%
\begin{tabular}{ccc|c}
$X_1$ & $X_2$ & $X_4$ & $\psi_1$ \\ \hline
0 & 0 & 0 & $v_1$ \\
\rowcolor{gray!50!white}
0 & 0 & 1 & $v_1$ \\
0 & 1 & 0 & $v_2$ \\
\rowcolor{gray!50!white}
0 & 1 & 1 & $v_1$ \\
1 & 0 & 0 & $v_3$ \\
\rowcolor{gray!50!white}
1 & 0 & 1 & $\infty$ \\
1 & 1 & 0 & $v_3$ \\
\rowcolor{gray!50!white}
1 & 1 & 1 & $\infty$
\end{tabular}};

\node at (9,0) (t3) {%
\begin{tabular}{cccc|c}
$X_1$ & $X_2$ & $X_3$ & $X_4$ & $\psi_{1\otimes 2}$ \\ \hline
0   & 0   & 0   & 0   & $v_1\otimes v_3$ \\
\rowcolor{gray!50!white}
0   & 0   & 0   & 1   & $v_1\otimes v_7$ \\
0   & 0   & 1   & 0   & $v_1\otimes v_3$ \\
\rowcolor{gray!50!white}
0   & 0   & 1   & 1   & $v_1\otimes v_1$ \\
0   & 1   & 0   & 0   & $v_2\otimes v_3$ \\
\rowcolor{gray!50!white}
0   & 1   & 0   & 1   & $v_1\otimes v_7$ \\
0   & 1   & 1   & 0   & $v_2\otimes v_3$ \\
\multicolumn{4}{c|}{\scriptsize$\vdots$} & \scriptsize$\vdots$
\end{tabular}};

\node at ($(t1.east)!0.5!(t3.west)$) (t2) {%
\begin{tabular}{cc|c}
$X_3$ & $X_4$ & $\psi_2$ \\ \hline
0 & 0 & $v_3$ \\
\rowcolor{gray!50!white}
0 & 1 & $v_7$ \\
1 & 0 & $v_3$ \\
\rowcolor{gray!50!white}
1 & 1 & $v_1$ \\
\end{tabular}};

\node at ($(t1.east)!0.5!(t2.west)$) {$\otimes$};

\node at ($(t2.east)!0.5!(t3.west)$) {$=$};

\end{tikzpicture}
\caption{\label{fig:sum}An example of the $\otimes$ operation.}
\end{figure}
\addtolength{\tabcolsep}{\reducecol}

The result of the $\otimes$ operation is a new function whose scope is the union of the scopes of the input functions, and in which the value of each variable assignment is the $\otimes\in\{\cdot,+\}$
of the values of the corresponding assignments (i.e., with the same assignments of the corresponding variables) in the input functions.
For example, the assignment $\tuple{X_1=0,X_2=1,X_3=1,X_4=0}$ in the result table corresponds to $\tuple{X_1=0,X_2=1,X_4=0}$ and $\tuple{X_3=1,X_4=0}$ in the input tables, hence its value is $v_2\otimes v_3$.
The $\otimes$ operation is closely related to the \emph{inner join} of relational algebra.

To efficiently implement $\dafsa(\psi_1)\otimes\dafsa(\psi_2)$ we will make use of the \emph{intersection} operation on automata \cite{DBLP:books/aw/HopcroftU79}.
Intuitively, the intersection of two automata accepting respectively $L_1$ and $L_2$ is an automaton that accepts $L_1\cap L_2$, i.e., all the strings appearing both in $L_1$ and $L_2$.
In our case, we will exploit the intersection operation to identify all the corresponding variable assignments in $\dafsa(\psi_1)$ and $\dafsa(\psi_2)$.
To make this possible, we first have to make sure that both functions have the same scope, so that corresponding levels in $\dafsa(\psi_1)$ and $\dafsa(\psi_2)$ correspond to the same variables.
We achieve this by means of the \textsc{AddLevels} operation.
\begin{definition}
Given two functions $\dafsa(\psi_1)$ and $\dafsa(\psi_2)$, the \textsc{AddLevels} operation inserts (i) one or more levels labeled with $\ast$ in each \gls{dafsa} and (ii) one or more variables in the respective scopes, in a way that the resulting scope is $var(\psi_1)\cup var(\psi_2)$.
Each level and variable is added so as to maintain the scope ordered \wrt{} the variable ordering.
\end{definition}

\begin{figure}[b]
\centering
\begin{tikzpicture}[xscale=0.36,yscale=0.75]

\tikzstyle{dafsastate}=[state,inner sep=0,minimum width=3mm,minimum height=3mm]
\def\v{6mm}
 
\node[fill=blue!20!white,draw,minimum width=\v,minimum height=\v] at (0,0) (tb) {$v_2$};
\node[draw,minimum width=\v,minimum height=\v] at ($ (tb) + (0,1.5) $) (ta) {$v_1$};
\node[fill=green!20!white,draw,minimum width=\v,minimum height=\v] at ($ (tb) - (0,1.5) $) (tc) {$v_3$};
\node[minimum width=\v,minimum height=\v] at ($ (ta) + (0,1.25) $) (tpsi) {$\psi_1$};


\node[dafsastate] at ($(ta) + (2,0)$) (ta1) {};
\node[dafsastate] at ($(ta1) + (2,0)$) (ta2) {};
\node[dafsastate] at ($(ta2) + (2,0.35)$) (ta3a) {};
\node[dafsastate] at ($(ta2) + (2,-0.35)$) (ta3b) {};
\node[dafsastate] at ($(ta3a) + (2,0)$) (ta5a) {};
\node[dafsastate] at ($(ta3b) + (2,0)$) (ta5b) {};
\node[dafsastate,accepting] at ($(ta2) + (6,0)$) (ta4) {};

\path[-{Stealth}]
(ta) edge [above] node [align=center] {} (ta1)
(ta1) edge [above] node [align=center] {0} (ta2)
(ta2) edge [above,sloped] node [align=center] {0} (ta3a);

\path[-{Stealth}]
(ta2) edge [below,sloped] node [align=center] {1} (ta3b);

\path[-{Stealth},densely dotted]
(ta3a) edge [above,sloped] node [align=center] {$\ast$} (ta5a);

\path[-{Stealth},densely dotted]
(ta3b) edge [above,sloped] node [align=center] {$\ast$} (ta5b);

\path[-{Stealth}]
(ta5a) edge [above,sloped] node [align=center] {$\ast$} (ta4);

\path[-{Stealth}]
(ta5b) edge [below,sloped] node [align=center] {1} (ta4);


\node[dafsastate] at ($(tb) + (2,0)$) (tb1) {};
\node[dafsastate] at ($(tb1) + (2,0)$) (tb2) {};
\node[dafsastate] at ($(tb2) + (2,0)$) (tb3) {};
\node[dafsastate] at ($(tb3) + (2,0)$) (tb5) {};
\node[dafsastate,accepting] at ($(tb5) + (2,0)$) (tb4) {};

\path[-{Stealth}]
(tb) edge [above] node [align=center] {} (tb1)
(tb1) edge [above] node [align=center] {0} (tb2)
(tb2) edge [above] node [align=center] {1} (tb3);

\path[-{Stealth},densely dotted]
(tb3) edge [above] node [align=center] {$\ast$} (tb5);

\path[-{Stealth}]
(tb5) edge [above] node [align=center] {0} (tb4);


\node[dafsastate] at ($(tc) + (2,0)$) (tc1) {};
\node[dafsastate] at ($(tc1) + (2,0)$) (tc2) {};
\node[dafsastate] at ($(tc2) + (2,0)$) (tc3) {};
\node[dafsastate] at ($(tc3) + (2,0)$) (tc5) {};
\node[dafsastate,accepting] at ($(tc5) + (2,0)$) (tc4) {};

\path[-{Stealth}]
(tc) edge [above] node [align=center] {} (tc1)
(tc1) edge [above] node [align=center] {1} (tc2)
(tc2) edge [above] node [align=center] {$\ast$} (tc3);

\path[-{Stealth},densely dotted]
(tc3) edge [above] node [align=center] {$\ast$} (tc5);

\path[-{Stealth}]
(tc5) edge [above] node [align=center] {0} (tc4);


\node at ($(tb1)!0.5!(tb2)$) (x1) {};
\node at (x1 |- tpsi) (X1) {$X_1$};

\node at ($(tb2)!0.5!(tb3)$) (x2) {};
\node at (x2 |- tpsi) (X2) {$X_2$};

\node at ($(tb3)!0.5!(tb5)$) (x3) {};
\node at (x3 |- tpsi) (X3) {$X_3^+$};

\node at ($(tb5)!0.5!(tb4)$) (x4) {};
\node at (x4 |- tpsi) (X4) {$X_4$};


\node[fill=green!20!white,draw,minimum width=\v,minimum height=\v] at ($(tb) + (12.5,0)$) (sb)  {$v_3$};
\node[draw,minimum width=\v,minimum height=\v] at ($ (sb) + (0,1.5) $) (sa) {$v_1$};
\node[fill=orange!20!white,draw,minimum width=\v,minimum height=\v] at ($ (sb) - (0,1.5) $) (sc) {$v_7$};
\node[minimum width=\v,minimum height=\v] at ($ (sa) + (0,1.25) $) (spsi) {$\psi_2$};


\node[dafsastate] at ($(sa) + (2,0)$) (sa1) {};
\node[dafsastate] at ($(sa1) + (2,0)$) (sa2) {};
\node[dafsastate] at ($(sa2) + (2,0)$) (sa3) {};
\node[dafsastate] at ($(sa3) + (2,0)$) (sa4) {};
\node[dafsastate,accepting] at ($(sa4) + (2,0)$) (sa5) {};

\path[-{Stealth},densely dotted]
(sa) edge [above] node [align=center] {} (sa1)
(sa1) edge [above] node [align=center] {$\ast$} (sa2)
(sa2) edge [above] node [align=center] {$\ast$} (sa3);

\path[-{Stealth}]
(sa3) edge [above] node [align=center] {1} (sa4)
(sa4) edge [above] node [align=center] {1} (sa5);


\node[dafsastate] at ($(sb) + (2,0)$) (sb1) {};
\node[dafsastate] at ($(sb1) + (2,0)$) (sb2) {};
\node[dafsastate] at ($(sb2) + (2,0)$) (sb3) {};
\node[dafsastate] at ($(sb3) + (2,0)$) (sb4) {};
\node[dafsastate,accepting] at ($(sb4) + (2,0)$) (sb5) {};

\path[-{Stealth},densely dotted]
(sb) edge [above] node [align=center] {} (sb1)
(sb1) edge [above] node [align=center] {$\ast$} (sb2)
(sb2) edge [above] node [align=center] {$\ast$} (sb3);

\path[-{Stealth}]
(sb3) edge [above] node [align=center] {$\ast$} (sb4)
(sb4) edge [above] node [align=center] {1} (sb5);


\node[dafsastate] at ($(sc) + (2,0)$) (sc1) {};
\node[dafsastate] at ($(sc1) + (2,0)$) (sc2) {};
\node[dafsastate] at ($(sc2) + (2,0)$) (sc3) {};
\node[dafsastate] at ($(sc3) + (2,0)$) (sc4) {};
\node[dafsastate,accepting] at ($(sc4) + (2,0)$) (sc5) {};

\path[-{Stealth},densely dotted]
(sc) edge [above] node [align=center] {} (sc1)
(sc1) edge [above] node [align=center] {$\ast$} (sc2)
(sc2) edge [above] node [align=center] {$\ast$} (sc3);

\path[-{Stealth}]
(sc3) edge [above] node [align=center] {0} (sc4)
(sc4) edge [above] node [align=center] {1} (sc5);


\node at ($(sb1)!0.5!(sb2)$) (x1) {};
\node at (x1 |- spsi) (X1) {$X_1^+$};

\node at ($(sb2)!0.5!(sb3)$) (x2) {};
\node at (x2 |- spsi) (X2) {$X_2^+$};

\node at ($(sb3)!0.5!(sb4)$) (x3) {};
\node at (x3 |- spsi) (X3) {$X_3$};

\node at ($(sb4)!0.5!(sb5)$) (x4) {};
\node at (x4 |- spsi) (X4) {$X_4$};

\end{tikzpicture}
\caption{\label{fig:add_level}The result of the \textsc{AddLevels} operation on $\dafsa(\psi_1)$ and $\dafsa(\psi_2)$, where $\psi_1$ and $\psi_2$ are the tables in Figure~\ref{fig:sum}. Added levels and variables are denoted with dotted lines and $^+$ superscript.}
\end{figure}
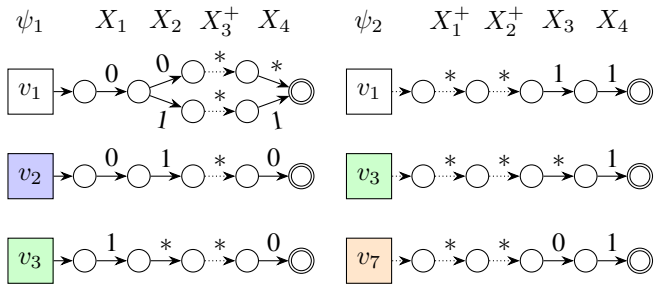

Figure~\ref{fig:add_level} shows an example of \textsc{AddLevels}.
The operation of adding one level to a \gls{dafsa} $\delta$ has a linear complexity \wrt{} the number of paths in $\delta$.
This has to be executed $\card{\dafsa(\psi_1)}\!\cdot\!\card{var(\psi_2)\!\setminus\!var(\psi_1)}\!+\!\card{\dafsa(\psi_2)}\!\cdot\!\card{var(\psi_1)\!\setminus\!var(\psi_2)}$ times.

Our \gls{dafsa}-based $\otimes$ operation is implemented by Algorithm~\ref{alg:dafsa_sum}.
Intuitively, for each couple of values $(v_i,v_j)$, where $v_i$ and $v_j$ are values in $\dafsa(\psi_1)$ and $\dafsa(\psi_2)$ respectively, we compute the variable assignments associated to their $\otimes$ by computing the intersection $\delta_i\cap \delta_j$ between the corresponding \gls{dafsa} $\delta_i$ and $\delta_j$.
The result is then associated to the value $v_i \otimes v_j$ in the output function.
We maintain only one entry for each value $v_i \otimes v_j$ (see Remark~\ref{rem:float} in this respect) by accumulating (i.e., taking the union of) all the \gls{dafsa} that are associated to the same value (Line~\ref{line:union}).

Union and intersection over \gls{dafsa} have a time complexity of $\bigO{nm}$~\cite{han2008state}, where $n$ and $m$ are the number of states of the input automata.
Depending on their implementations, such operations may not directly produce a minimal \gls{dafsa}.
Nonetheless, \gls{dafsa} can be minimized in linear time \wrt{} the number of states with the algorithm by \authorref{Bubenzer}{DBLP:conf/stringology/Bubenzer11}.

\begin{algorithm}[t]\caption{$\dafsa(\psi_1)\otimes\dafsa(\psi_2)$}\label{alg:dafsa_sum}
\begin{algorithmic}[1]
\STATE $(\dafsa(\psi_1)',\dafsa(\psi_2)')=\textsc{AddLevels}(\dafsa(\psi_1),\dafsa(\psi_2))$.
\FORALL{$(v_i,\delta_i)\in\dafsa(\psi_1)', (v_j,\delta_j)\in\dafsa(\psi_2)'$}
\IF{$\exists(v_i \otimes v_j,\delta_{k})\in\dafsa(\psi_1)\otimes\dafsa(\psi_2)$}
\STATE $\delta_{k}=\delta_{k}\cup (\delta_i\cap \delta_j)$.\label{line:union}
\ELSE
\STATE Add $\{(v_i \otimes v_j,\delta_i\cap \delta_j)\}$ to $\dafsa(\psi_1)\otimes\dafsa(\psi_2)$.
\ENDIF
\ENDFOR
\RETURN $\dafsa(\psi_1)\otimes\dafsa(\psi_2)$.
\end{algorithmic}
\end{algorithm}

\subsection{A \gls{dafsa}-Based Version of $\Downarrowmath$}

The $\Downarrowmath\in\{\max,\min\}$ operation effectively realizes \emph{variable elimination} within the \gls{be} algorithm.
Specifically, $\Downarrowmath_{X_i}\psi$ removes $X_i$ from the scope of $\psi$, and, from all the rows that possibly have equal variable assignments as a result of the elimination of the column associated to $X_i$, it only maintains the one with the $\max$ (in the case of \gls{map}, or $\min$ in the case of optimization) value.
Like $\otimes$, $\Downarrowmath$ is also related to a relational algebra operation, i.e., the \emph{project} operation.
In terms of SQL, $\Downarrowmath_{X_i}\psi$ is equivalent to \texttt{SELECT} $var(\psi)\setminus X_i,\max(\psi(\cdot))$ \texttt{FROM} $\psi$ \texttt{GROUP} \texttt{BY} $var(\psi)\setminus X_i$, in the case of $\max$.

We realize the elimination of the column associated to $X_i$ with the \textsc{RemoveLevel} operation, which can be intuitively thought of as the inverse of \textsc{AddLevels}.
\textsc{Remove} $\textsc{Level}(\dafsa(\psi),X_i)$ removes $X_i$ from the scope of $\dafsa(\psi)$ and collapses all the edges at the level associated to $X_i$ from all the \gls{dafsa} in $\dafsa(\psi)$.
\textsc{RemoveLevel} has similar computational properties \wrt{} \textsc{AddLevels}, so we do not repeat the discussion due to space limitations.
Notice that the \textsc{RemoveLevel} operation could result in a \emph{non-deterministic} automaton if the removal happens in correspondence of a branching.
Our approach takes this into account by using a determinization algorithm \cite{DBLP:books/aw/HopcroftU79}.

In general, determinising an automaton could produce a growth (up to exponential, in the worst case) of the number of states. 
On the other hand, in all our experiments such a worst-case never happens and the growth factor is, on average, only around 10\%.
Results confirm that such a growth does not affect the overall performance of our approach, which is able to outperform the competitors as described in Section~\ref{sec:exp}.

We then implement the maximization (resp. minimization) of the values as follows.
Without loss of generality, we assume that the values $v_1,\ldots,v_{\card{\dafsa(\psi)}}$ are in decreasing (resp. increasing) order.
For each $(v_i,\delta_i)\in\dafsa(\psi)$, we subtract from $\delta_i$ all $\delta_j$ such that $v_j$ precedes $v_i$ in the above-mentioned ordering (i.e., $v_j\geq v_i$, resp. $\leq$).
In this way, we remove all duplicate variable assignments and we ensure that each assignment is \emph{only} associated to the maximum (resp. minimum) value, correctly implementing the $\Downarrowmath$ operation.
Subtraction over \gls{dafsa} has a time complexity of $\bigO{nm}$~\cite{han2008state}, where $n$ and $m$ are the number of states of the input automata.
Algorithm \ref{alg:dafsa_min} details our $\Downarrowmath$ implementation.

\begin{algorithm}[t]\caption{$\Downarrowmath_{X_i}\dafsa(\psi)$}\label{alg:dafsa_min}
\begin{algorithmic}[1]
\STATE $\dafsa(\psi)'=\textsc{RemoveLevel}(\dafsa(\psi),X_i)$.
\FORALL{$(v_i,\delta_i)\in\dafsa(\psi)'$ with decr. (resp. incr.) $v_i$}
\STATE $\delta_i=\delta_i\setminus\delta_{prec}$.
\STATE $\delta_{prec}=\delta_{prec}\cup\delta_i$.
\ENDFOR
\RETURN $\dafsa(\psi)'$.
\end{algorithmic}
\end{algorithm}

We remark that both our versions of $\otimes$ and $\Downarrowmath$ entirely operate on our concise representation, never expanding any function to a full table.
We directly employ our $\otimes$ and $\Downarrowmath$ operations within Algorithm~\ref{alg:be}.
\glsreset{fabe}
We call our \gls{dafsa}-based version of \gls{be} \gls{fabe}.
Since the results of our $\otimes$ and $\Downarrowmath$ operations are equivalent to the original ones, it follows that, as \gls{be}, \gls{fabe} is also an exact algorithm.
Finally, we remark that our $\otimes$ and $\Downarrowmath$ operations can directly be used within the approximated version of \gls{be}, i.e., \gls{mbe} \cite{DBLP:conf/ijcai/Dechter97}.

\newcommand{\timeout}{\color{red}{$>$ 2 h}}
\sisetup{detect-all,table-format=4.2}

\begin{table*}[h!]
    \setlength{\tabcolsep}{2.9pt}
    \centering
    \footnotesize
    \begin{tabular}{cSSSSSSSSSSSS}
        \toprule
        \texttt{protein}    & \texttt{1duw}     & \texttt{1hcz}     & \texttt{1fny}     & \texttt{2hft}     & \texttt{1ad2}     & \texttt{1atg}
                            & \texttt{1qre}     & \texttt{1qhv}     & \texttt{1pbv}     & \texttt{1g3p}     & \texttt{2fcb}     & \texttt{1euo}     \\
        \midrule
        \gls{fabe}          & \bfseries 21.36   & \bfseries 10.33   & \bfseries 6.60    & 322.33            & \bfseries 25.28   & 3.28
                            & \bfseries 3.47    & \bfseries 16.54   & 234.70            & \bfseries 1.21    & \bfseries 3.80    & \bfseries 39.81   \\
        \gls{rbfaoo}        & \timeout{}        & 749.39            & \timeout{}        & 1765.22           & 1654.75           & 1697.87
                            & 734.85            & \timeout{}        & 1543.11           & 5080.64           & 1677.44           & \timeout{}        \\
        \gls{smp}           & \timeout{}        & \timeout{}        & 2036.29           & 6569.95           & \timeout{}        & 4098.89
                            & 1721.50           & 4376.94           & \timeout{}        & 580.33            & 6780.26           & 1374.51           \\
        \texttt{toulbar}    & 25.89             & 13.12             & 8.09              & \bfseries 0.02    & 25.75             & \bfseries 0.04
                            & 6.87              & 23.08             & \bfseries 0.05    & 2.05              & 3.97              & 57.08             \\
        \midrule
        \texttt{pedigree}   & \texttt{25}       & \texttt{30}       & \texttt{39}       & \texttt{18}       & \texttt{31}       & \texttt{34}
                            & \texttt{51}       & \texttt{9}        & \texttt{13}       & \texttt{7}        & \texttt{41}       & \texttt{37}       \\
        \midrule
        \gls{fabe}          & 28.82             & \bfseries 7.23    & \bfseries 3.21    & \bfseries 7.42    & \bfseries 910.46  & \bfseries 8.83
                            & \bfseries 132.92  & 473.94            & \bfseries 519.08  & \bfseries 21.79   & \bfseries 24.06   & \bfseries 6.73    \\
        \gls{rbfaoo}        & \bfseries 6.32    & 61.34             & 22.46             & 20.11             & \timeout{}        & \timeout{}
                            & \timeout{}        & \bfseries 100.19  & \timeout{}        & 323.77            & \timeout{}        & 146.23            \\
        \gls{smp}           & 197.89            & 40.86             & 17.06             & 40.89             & 5881.66           & 60.78
                            & 789.65            & 2040.32           & 2011.26           & 136.11            & 151.40            & 27.91             \\
        \texttt{toulbar}    & 5962.05           & 581.43            & 17.43             & 166.29            & 1505.02           & 9.59
                            & \timeout{}        & \timeout{}        & 623.89            & 24.23             & 727.16            & 7.89              \\
        \midrule
        \texttt{grid}       & \prob{90}{26}{5}  & \prob{90}{25}{5}  & \prob{90}{24}{5}  & \prob{75}{23}{5}  & \prob{90}{23}{5}  & \prob{75}{22}{5}
                            & \prob{90}{22}{5}  & \prob{75}{21}{5}  & \prob{90}{21}{5}  & \prob{50}{20}{5}  & \prob{75}{20}{5}  & \prob{90}{20}{5}  \\
        \midrule
        \gls{fabe}          & 3192.15           & \timeout{}        & 5112.09           & \timeout{}        & 508.75            & \timeout{}
                            & 4883.60           & \timeout{}        & 142.23            & \timeout{}        & 831.80            & 267.50            \\
        \gls{rbfaoo}        & 925.47            & 902.19            & 1758.19           & 791.70            & 158.17            & 816.87
                            & 20.37             & 4.71              & 8.22              & \bfseries 163.44  & 19.01             & 10.27             \\
        \gls{smp}           & \timeout{}        & \timeout{}        & \timeout{}        & \timeout{}        & 2453.45           & \timeout{}
                            & \timeout{}        & \timeout{}        & 802.17            & \timeout{}        & 3711.20           & 946.07            \\
        \texttt{toulbar}    & \bfseries 0.04    & \bfseries 0.07    & \bfseries 0.02    & \bfseries 1.68    & \bfseries 0.03    & \bfseries 1.44
                            & \bfseries 0.02    & \bfseries 0.20    & \bfseries 0.02    & 1030.17           & \bfseries 0.10    & \bfseries 0.01    \\
        \bottomrule
    \end{tabular}
    \caption{\label{tab:map}Runtime results (in seconds) on 12 largest \gls{map} instances.}
\end{table*}

\begin{table*}
    \centering
    \footnotesize
    \setlength{\tabcolsep}{20.2pt}
    \def\uns#1#2{~(#1\%,~#2\%)}
    \def\sp#1#2{#1 speed-up \wrt{} #2}
    \sisetup{
        table-figures-integer = 3,
        table-figures-decimal = 1,
        round-mode = places,
        round-precision = 1,
        table-space-text-post = {xxxxxxxxxxx},
    }
    \begin{tabular}{cSSSSSS}
        \toprule
        Dataset                             & \texttt{protein}          & \texttt{pedigree}         & \texttt{grid}             \\
        \midrule                                                                                                                                                                        
        Average Redundancy                  & {96\%}                    & {85\%}                    & {64\%}                    \\
        \sp{\gls{fabe}}{\gls{rbfaoo}}       & 58.6 \uns{1}{11}          & 5.5 \uns{0}{32}           & 0.1 \uns{43}{0}           \\
        \sp{\gls{fabe}}{\gls{smp}}          & 1006.5 \uns{1}{38}        & 6.8 \uns{0}{5}            & 4.0 \uns{43}{70}          \\
        \sp{\gls{sprbfaoo}}{\gls{rbfaoo}}   & {$\sim$7}                 & {$\sim$7}                 & {$\sim$5}                 \\
        \sp{\gls{fabe}}{\texttt{toulbar}}   & 1.0 \uns{1}{0}            & 20.6 \uns{0}{23}          & 0.0 \uns{43}{0}           \\
        \midrule
        Dataset                             & \texttt{spot5}            & \texttt{mastermind}       & \texttt{iscas89}          \\
        \midrule                                                                                                                                                                        
        Average Redundancy                  & {85\%}                    & {85\%}                    & {87\%}                    \\
        \sp{\gls{fabe}}{\gls{rbfaoo}}       & 36.9 \uns{0}{0}           & 6.2 \uns{0}{0}            & 0.4 \uns{0}{0}            \\
        \sp{\gls{fabe}}{\gls{smp}}          & 5.8 \uns{0}{0}            & 2.5 \uns{0}{0}            & 5.1 \uns{0}{0}            \\
        \sp{\gls{fabe}}{\texttt{toulbar}}   & 10615.5 \uns{0}{50}       & 0.4 \uns{0}{0}            & 0.1 \uns{0}{0}            \\
        \sp{\gls{fabe}}{CPLEX}              & 2.0 \uns{0}{0}            & 7.2 \uns{0}{0}            & 0.8 \uns{0}{0}            \\
        \sp{\gls{fabe}}{GUROBI}             & 1.2 \uns{0}{0}            & 4.4 \uns{0}{0}            & 0.8 \uns{0}{0}            \\
        \bottomrule
    \end{tabular}
    \caption{\label{tab:cumulative}Average speed-up results for \gls{map} (top) and \gls{wcsp} (bottom) instances. For \gls{sprbfaoo} we report the same speed-up values reported by the authors \protect\cite{DBLP:conf/nips/KishimotoMB15}. Values in parentheses indicate the percentages of instances unsolved by first and second approach.}
\end{table*}

\section{Experiments}\label{sec:exp}

We evaluate \gls{fabe} on standard benchmark datasets\footnote{Online at: \url{www.ics.uci.edu/~dechter/softwares/benchmarks}.} for exact \gls{map} inference (i.e., \texttt{protein}, \texttt{pedigree}, \texttt{grid})~\cite{DBLP:conf/uai/Kishomoto014,DBLP:conf/nips/KishimotoMB15} and \gls{wcsp} (i.e., \texttt{spot5}, \texttt{mastermind}, \texttt{iscas89})~\cite{DBLP:journals/ai/MarinescuD09}, comparing it with several state of the art approaches.
For \gls{map}, we consider \gls{rbfaoo} \cite{DBLP:conf/uai/Kishomoto014} as our main competitor since it has been empirically shown to be superior to other sequential algorithms for exact \gls{map} inference, namely \gls{aobb} \cite{DBLP:journals/ai/MarinescuD09} and \gls{aobf} \cite{DBLP:conf/aaai/MarinescuD07} (see Section \ref{sec:relwork}).
We cannot directly compare against the \emph{parallel} version of \gls{rbfaoo}, i.e., \gls{sprbfaoo} \cite{DBLP:conf/nips/KishimotoMB15}, because its implementation has not been made public.
We discarded the option of re-implementing \gls{sprbfaoo}, as it would have probably led to an unfair comparison due to a sub-optimal implementation.
Nonetheless, since \gls{rbfaoo} is also used as a baseline for speed-up calculation in \cite{DBLP:conf/nips/KishimotoMB15}, in Table \ref{tab:cumulative} we compare our values of speed-up with the ones reported for \gls{sprbfaoo} by its authors.
As a second competitor we consider \texttt{toulbar} \cite{hurley2016multi}, a widely used open-source solver used for exact optimization.
We also compare \gls{fabe} against the \gls{smp} approach by \authorref{Gogate and Domingos}{10.5555/3023638.3023664} (see associated discussion in Section~\ref{sec:relwork}).
Since \gls{smp} relies on \glspl{add} (which cannot represent non-binary variables natively), we encode non-binary variables using \emph{one-hot} encoding, following a standard practice.
We do not show results comparing \gls{fabe} against the standard version of \gls{be} with tabular functions \cite{DBLP:series/synthesis/2013Dechter}, since the latter runs out of memory on most of the instances. 
For \glspl{wcsp} we also compare against CPLEX and GUROBI, two off-the-shelf solvers for constrained optimization.

\gls{fabe}'s reported runtimes also include the compilation of automata.
Since both \gls{fabe} and \gls{rbfaoo} require an ordering $d$ (see Section \ref{sec:back}), we consider this as pre-processing and we do not include its runtime in the results (also because it is negligible \wrt{} the solution phase).
For each instance, we compute $d$ using a weighted \textsc{Min-Fill} heuristic~\cite{DBLP:series/synthesis/2013Dechter}, and we use the same $d$ for both algorithms. 

We execute \gls{rbfaoo} with the parameters detailed in authors' previous work~\cite{DBLP:conf/uai/Kishomoto014,DBLP:conf/nips/KishimotoMB15}, including cache size and $i$ parameter.
Every other approach is executed with default parameter values.
Following \cite{DBLP:conf/nips/KishimotoMB15}, we set a time limit of 2 hours.
We exclude from our analysis all instances that could not be solved by any algorithm in the considered time limit.
\gls{fabe} and \gls{smp} are implemented in C++.\footnote{Code available at \url{https://github.com/filippobistaffa/FABE}.}
We employ the implementations of \gls{rbfaoo} and \texttt{toulbar} provided by the authors.
All experiments have been run on a cluster whose computing nodes have 2.50GHz CPUs and 384 GBytes of RAM.
As for Remark~\ref{rem:float}, for \gls{fabe} we consider $\epsilon\!=\!10^{-10}$.

Given the large number of instances in \gls{map} datasets, in Table~\ref{tab:map} we report the runtimes on the 12 largest instances \wrt{} the number of variables.
In Table \ref{tab:cumulative} we report the aggregated results of the speed-ups achieved by \gls{fabe} \wrt{} other approaches, calculated considering the instances where both algorithms terminate within the time limit.

\subsection{Summary of Results}
Results confirm that \gls{fabe}'s performance depends on the degree of redundancy. 
\gls{fabe} obtains good performance on the \texttt{protein} and \texttt{pedigree} datasets, achieving speed-ups of $\sim$1--2 orders of magnitude, and solving a total of 34 instances that \gls{rbfaoo} could not solve.
We also observe that \texttt{toulbar} is superior on a handful of \texttt{protein} instances and on the \texttt{grid} dataset, which is characterized by low redundancy.
Results also show that, despite not employing parallelism, \gls{fabe}'s speed-up on the \texttt{protein} dataset is much higher than the one reported for \gls{sprbfaoo}, while it is comparable on the \texttt{pedigree} datasets.

As for \glspl{wcsp} (detailed results in Table~\ref{tab:wcsp}), \gls{fabe} outperforms all competitors on the \texttt{spot5} dataset, notably achieving a speed-up of 5 orders of magnitude \wrt{} \texttt{toulbar}.
On the \texttt{mastermind} dataset, \gls{fabe} is comparable with \texttt{toulbar} (since both compute solutions in tenths of seconds) except for \prob{3}{8}{5} and \prob{10}{8}{3} instances, but it is, on average, better than every other competitor.
We also observe that \texttt{toulbar} is clearly superior on the \texttt{iscas89} dataset.

Finally, \gls{fabe} consistently outperforms \gls{smp} using one-hot encoding, confirming that the use of additional encodings (required by the presence of non-binary variables that cannot be represented by \glspl{add}) introduces a significant overhead compared to our representation using \gls{dafsa}, which can natively represent non-binary variables.
Such an impact is more pronounced on datasets with larger variable domains.
Indeed, \gls{fabe} obtains a speed-up of 3 orders of magnitude on the \texttt{protein} dataset, where variables reach a domain of 81.

\section{Conclusions}
\label{sec:concl}

We proposed \gls{fabe}, an algorithm for exact \gls{map} and constrained optimization that exploits our function representation based on \gls{dafsa}.
Results obtained following an established methodology confirm the efficacy of our representation.

Future research directions include extending \gls{fabe} to marginal inference tasks and integrating \gls{fabe} to compute the initial heuristic for AND/OR search algorithms, which, at the moment, use the table-based implementation of \gls{be}.
The computation of such an heuristic represents a bottleneck for high values of $i$, as acknowledged in \cite{DBLP:conf/nips/KishimotoMB15}.
A faster version of \gls{mbe} could represent an important contribution, allowing one to employ more precise heuristics and, thus, improving the performance of the AND/OR search.

\begin{table}[H]
    \setlength{\tabcolsep}{2pt}
    \centering
    \footnotesize
    \sisetup{
        round-mode = places,
        round-precision = 2
    }
    \begin{tabular}{cSSSSSS}
        \toprule
        \texttt{spot5}      & \texttt{42b}      & \texttt{505b}     & \texttt{408b}     & \texttt{29}       & \texttt{503}      & \texttt{54}       \\
        \midrule
        \gls{fabe}          & \bfseries 0.26    & \bfseries 0.26    & \bfseries 0.29    & \bfseries 0.09    & \bfseries 0.05    & \bfseries 0.07    \\
        \gls{rbfaoo}        & 13.37             & 10.27             & 9.97              & 5.61              & 1.37              & 1.36              \\
        \gls{smp}           & 1.62              & 1.80              & 1.60              & 0.72              & 0.27              & 0.30              \\
        CPLEX               & 0.35              & 0.34              & 0.39              & 0.37              & 0.10              & 0.18              \\
        GUROBI              & 0.38              & 0.40              & 0.41              & 0.13              & 0.25              & 0.21              \\
        \texttt{toulbar}    & \timeout{}        & \timeout{}        & \timeout{}        & 0.10              & 1957.80           & 0.09              \\
        \midrule
        \texttt{master.}    & \prob{3}{8}{5}    & \prob{10}{8}{3}   & \prob{4}{8}{4}    & \prob{3}{8}{4}    & \prob{4}{8}{3}    & \prob{3}{8}{3}    \\  
        \midrule
        \gls{fabe}          & 247.27            & 69.30             & 0.36              & 0.22              & 0.10              & \bfseries 0.06    \\
        \gls{rbfaoo}        & 4.93              & 3.01              & 2.96              & 1.96              & 0.85              & 0.68              \\
        \gls{smp}           & 659.42            & 185.45            & 0.95              & 0.43              & 0.29              & 0.11              \\
        CPLEX               & 3.26              & 0.66              & 1.26              & 0.67              & 1.54              & 1.28              \\
        GUROBI              & 1.77              & 0.66              & 0.72              & 0.51              & 0.97              & 0.75              \\
        \texttt{toulbar}    & \bfseries 0.18    & \bfseries 0.09    & \bfseries 0.09    & \bfseries 0.12    & \bfseries 0.05    & \bfseries 0.06    \\
        \midrule
        \texttt{iscas89}    & \texttt{s1238}    & \texttt{c880}     & \texttt{s1196}    & \texttt{s953}     & \texttt{s1494}    & \texttt{s1488}    \\
        \midrule
        \gls{fabe}          & 38.50             & 25.36             & 73.54             & 286.43            & 1.42              & 1.12              \\
        \gls{rbfaoo}        & 1.47              & 1.17              & 0.61              & 0.54              & 0.41              & 0.39              \\
        \gls{smp}           & 229.64            & 146.43            & 410.96            & 1464.98           & 9.78              & 6.02              \\
        CPLEX               & 0.19              & 0.20              & 0.09              & 0.19              & 0.28              & 0.23              \\
        GUROBI              & 0.17              & 0.14              & 0.09              & 0.15              & 0.26              & 0.17              \\
        \texttt{toulbar}    & \bfseries 0.04    & \bfseries 0.06    & \bfseries 0.04    & \bfseries 0.04    & \bfseries 0.04    & \bfseries 0.07    \\
        \bottomrule
    \end{tabular}
    \caption{\label{tab:wcsp}Runtime results (in seconds) on \gls{wcsp} instances.}
\end{table}

\section*{Ethical Statement}

There are no ethical issues.

\section*{Acknowledgments}
The author was supported by the ``YOMA Operational Research'' project funded by the Botnar Foundation.
The author gratefully acknowledges the computer resources at Artemisa, funded by the European Union ERDF and Comunitat Valenciana as well as the technical support provided by the Instituto de Fisica Corpuscular, IFIC (CSIC-UV).

\bibliographystyle{named}
\bibliography{main}

\end{document}